\documentclass{article}
\usepackage{ijcai24}

\usepackage{times}
\usepackage{soul}
\usepackage{url}
\usepackage[hidelinks]{hyperref}
\usepackage[utf8]{inputenc}
\usepackage[small]{caption}
\usepackage{graphicx}
\usepackage{amsmath}
\usepackage{amsthm}
\usepackage{booktabs}
\usepackage{algorithm}
\usepackage{algorithmic}
\usepackage[switch]{lineno}
\usepackage{subcaption}
\usepackage{xcolor}

\title{Knowledge Distillation in Federated Learning: A Practical Guide}

\author{
Alessio Mora$^{1}$
\and
Irene Tenison$^2$\and
Paolo Bellavista$^1$\And
Irina Rish$^3$\\
\affiliations
$^1$University of Bologna\\
$^2$ Massachusetts Institute of Technology (MIT)\\
$^3$Mila, University of Montreal
\emails
alessio.mora@unibo.it,
itenison@mit.edu,
paolo.bellavista@unibo.it,
irina.rish@mila.quebec
}

\begin{document}

\maketitle

\begin{abstract}
    Federated Learning (FL) enables the training of Deep Learning models without centrally collecting possibly sensitive raw data. The most used algorithms for FL are parameter-averaging based schemes (e.g., Federated Averaging) that, however, have well known limits, i.e., model homogeneity, high communication cost, poor performance in presence of heterogeneous data distributions. Federated adaptations of regular Knowledge Distillation (KD) can solve or mitigate the weaknesses of parameter-averaging FL algorithms while possibly introducing other trade-offs. In this article, we originally present a focused review of the state-of-the-art KD-based algorithms specifically tailored for FL, by providing both a novel classification of the existing approaches and a detailed technical description of their pros, cons, and tradeoffs.
\end{abstract}

\section{Introduction}
Federated Learning (FL) has been proposed as an alternative to cloud-based Deep Learning (DL). This paradigm decouples the ability to train DL models from the need of harvesting raw data, alternating on-device computation and periodic communication \cite{mcmahan2016communication,bellavista2021decentralised}. During the learning process, only ephemeral and locally processed payloads need to be disclosed by the participants in the federation, making it harder to infer private information about the individuals.


Federated Averaging (FedAvg) represents the baseline algorithm for Federated Learning (FL) \cite{mcmahan2016communication}. In FedAvg, collaborative learning proceeds in synchronous rounds by leveraging a client-server paradigm. Participants (i.e., clients) iteratively exchange model updates and model weights with a central aggregator (i.e., the server) to collaboratively build a global model. Round by round, the server aggregates model updates by weighted average and distributes the new version of the global model.
However, parameter-averaging aggregation schemes, such as FedAvg, have well-known limits. Firstly, this class of algorithms implies model homogeneity among the federation, i.e., each client is constrained to use the same neural architecture since the server directly merges clients' updates (e.g., by weighted average). This may be an issue when the federation of learners consists of clients with heterogeneous hardware capabilities. Furthermore, exchanging model parameters and model updates have high communication cost, which scales with the number of model parameters -- even though a plethora of strategies (e.g., \cite{sattler2019robust}) have been proposed to improve the communication efficiency at the cost of global model performance. In addition, exchanging model parameters/updates exposes client to information leakage, and the server must know the architecture and structure of clients' model to broadcast the global parameters, possibly incurring in intellectual property issues (i.e., clients in the federation are unwilling to share the architecture they are using). Lastly, but not less important, when clients hold heterogeneous data, local models tend to diverge from each other during training and fine tune on private examples (i.e., client drift). As a consequence, directly aggregating model parameters/updates degrades the global model performance \cite{karimireddy2019scaffold}.

This article focuses on specifically reviewing the FL-oriented adaptations of regular Knowledge Distillation (KD) techniques that have been employed to alleviate the above mentioned  weaknesses of FL parameter-averaging aggregation schemes. Initially, KD-based strategies, also motivated by encouraging privacy properties \cite{papernot2016semi}, have been introduced to enable model heterogeneity and to reduce communication costs by exchanging model outputs and/or model-agnostic intermediate representations instead of directly transferring model parameters/updates. 
Then, a set of strategies have been proposed to enhance the aggregation step of FedAvg with a server-side ensemble distillation phase to enable model heterogeneity and/or improve model fusion in presence of heterogeneous data. 
Recently, two KD-based lines of work have focused on mitigating the phenomenon of client model drift -- which makes averaging-based aggregations inefficient -- either using regularization terms in clients' objective functions 
or leveraging globally learned data-free generator. 

To our knowledge, this is the first paper that provides a systematic categorization of KD-based methods tailored to address specific FL issues. The paper mainly makes the following contributions:
\begin{itemize}
    \item We propose a novel taxonomy of KD-based methods for FL, which can help researchers to better understand the potential and possible further applications of distillation-inspired methods.
    \item We present a detailed technical overview of the existing KD-based methods, of their primary scope, and of the rationale behind their design/implementation choices. In addition, we discuss their strengths and possible drawbacks.
    \item We outline the emerging use of KD in this domain, by highlighting promising directions for future research.
\end{itemize}

The paper is organized as follows. Section \ref{sec:background} introduces the fundamental of KD. Section \ref{sec:modelagnostic} presents FL algorithms that use KD to enable model heterogeneity. Section \ref{sec:dataagnostic} describes FL algorithms that use KD to mitigate the impact of data heterogeneity on global model performance. Finally, in Section \ref{sec:future_dir}, we concisely highlight the emerging use of KD for other relevant FL issues. 

\section{Background}
\label{sec:background}
\subsection{Knowledge Distillation}
Knowledge Distillation (KD) methods have been designed to transfer knowledge from a larger deep neural network, the \textit{teacher}, to a lightweight network, the \textit{student} \cite{hinton2015distilling}. In the simplest form of KD, the student model learns by mimicking the (pre-trained) teacher model's outputs on a proxy dataset, also called transfer set. If the transfer set is labeled, the student can be trained using a linear combination of two loss functions,
\begin{equation}
\label{eq:kd}
    \mathcal{L} = (1-\lambda) \mathcal{L}_{CE}(\tilde{y}^S, y) + \lambda \tau^2 \mathcal{L}_{KD}(\tilde{y}^S_{\tau}, \tilde{y}^T_{\tau})
\end{equation}
$\mathcal{L}_{CE}$ is the usual cross-entropy loss between the true label $y$ (e.g., hot encoded) and the class probabilities $\tilde{y}^S$ (i.e., soft targets) predicted by the student neural network. Soft targets $\tilde{y}_{\tau}$ are typically produced by applying a softmax layer to the logits $z_i$ so that $\tilde{y}(i) = \frac{exp(z_i/\tau)}{\sum_j exp(z_j/\tau)}$, where $z_i$ is the i-th value of logits vector $z$. The temperature $\tau$ controls the softness of the probability distribution. $\tilde{y}^S$ is computed with $\tau$ set to 1. $\lambda$ weights the impact of the two loss terms. $\mathcal{L}_{KD}$ is a general distillation loss that measures the distance between the distribution of student's soft targets ($\tilde{y}^S$) and the  the distribution of teacher's soft targets ($\tilde{y}^T$), e.g. via Kullback-Leibler (KL) divergence. Alternatively, $\mathcal{L}_{KD}$ could directly measure the error between student and teacher logits (e.g., mean squared error). We refer to \cite{gou2021knowledge} for taxonomy and recent progress in the KD area.

\subsection{Codistillation}
\label{subsec:codistillation}
Codistillation (CD) refers to an online version of distillation, which obviates the need of a pre-trained teacher in regular KD \cite{anil2018large,gou2021knowledge}. In fact, CD simultaneously trains $T$ copies of a model by adding a distillation term to the regular loss function of the jth model to mimic the average prediction of the other $T - 1$ models. In this way, each worker network sees the ensemble of the other models as a virtual teacher. For CD, the pre-trained teacher's soft targets in Eq. \ref{eq:kd} is replaced with the ensemble soft targets of $T-1$ workers.

In the original formulation of CD \cite{anil2018large}, (1) all the workers implement the same neural architecture, (2) all the workers use the same dataset for training and, most notably, (3) the distillation loss is employed during training before any model has fully converged. 

\subsection{Proposed Taxonomy and Classification}
\begin{figure}[t!]
\centering
\includegraphics[width=0.47\textwidth]{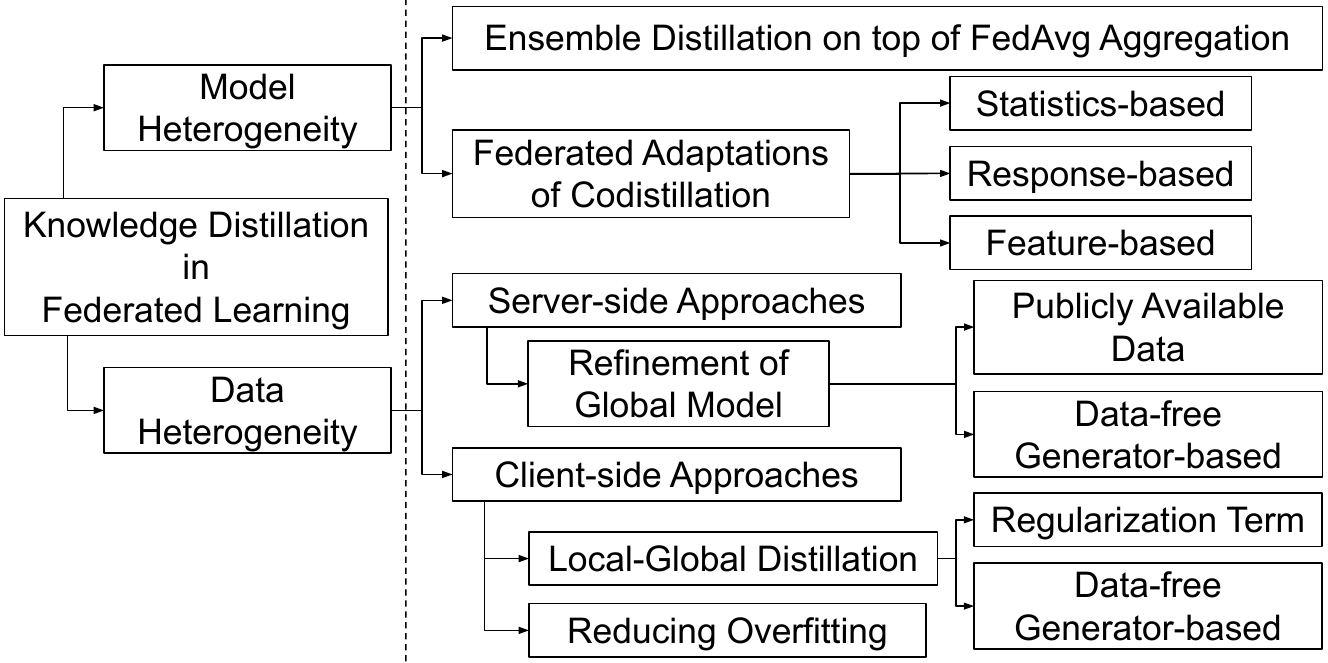}
\caption{Taxonomy of KD-based solutions for FL issues.}
\label{fig:roadmap}
\end{figure}

Figure \ref{fig:roadmap} illustrates our taxonomy of the possible KD-based mechanisms designed to enable model heterogeneity or to combat the effects of data heterogeneity. Table \ref{tab:large_comparison} lists the most relevant related work in the literature and surveyed in this paper, by classifying the proposed solutions according to their primary aim. For each solution, we detail the kind of per-round exchanged information, the need of auxiliary data, and the type of KD involved. For the latter, a regularizer-based approach uses KD to regularize local training. Generator-based mechanism leverages a generator model to assemble synthetic data and transfer knowledge as inductive biases. Digestion means that knowledge is absorbed by imitating teacher outputs on the same proxy data.


\begin{table*}[t]
\begin{center}
{
\footnotesize
\begin{tabular}{c c c c c c c}
\toprule
 & Purpose & \multicolumn{2}{c}{Exchanged information} & Auxiliary data & \multicolumn{2}{c}{KD approach} \\
 \cmidrule{3-4}\cmidrule{6-7}
 
 & & Upload & Download & & Client-side & Server-side \\
\midrule
FedDistill \cite{jeong2018communication} & MH, CE & $\overline{z}$ & $\overline{z}$ & data-free & regularizer & - \\
FedMD \cite{li2019fedmd} & MH, CE &$\tilde{y}_p$ & $\tilde{y}_p$ & labeled & digestion & - \\
Cronus \cite{chang2019cronus} & MH, CE & $\tilde{y}_p$ & $\tilde{y}_p$ & unlabeled & digestion & -  \\
DS-FL\cite{itahara2020distillation} & MH, CE & $\tilde{y}_p$ & $\tilde{y}_p$ & unlabeled & digestion & -\\
MATH \cite{hu2021mhat} & MH, CE & $\tilde{y}_p$ & $\tilde{y_p}$ & labeled & digestion & digestion\\
CFD \cite{sattler2021cfd} & MH, CE & $\tilde{y}_p$ & $\tilde{y}_p$ & unlabeled & digestion & digestion \\
FedGEMS \cite{cheng2021fedgems} &  MH, CE & $\tilde{y}_p$ & $\tilde{y}_p$ & labeled & digestion & digestion \\
FedAD \cite{gong2021ensemble} & MH, CE & $z_p$, $A_p$ & - & unlabeled & - & digestion \\
FedGKT \cite{he2020group} & MH, CE & $z, H, y$ & $z$ & data-free & regularizer & regularizer \\
FedDKC \cite{wu2022exploring} & MH, CE & $z, H, y$ & $z$ & data-free & regularizer & regularizer \\
FedDF \cite{lin2020ensemble} & MH, NIID & $w$ & $w$ & unlabeled & - & digestion \\
FedAUX \cite{sattler2021fedaux} & MH, NIID & $w$ & $w$ & unlabeled & - & digestion\\
FedBE \cite{chen2020fedbe} & NIID & $w$ & $w$ & unlabeled & - & digestion\\
FedFTG \cite{zhang2022fine} & NIID & $w, c$ & $w$ & data-free & - & generator + digestion \\
FedZKT \cite{zhang2021fedzkt} & NIID & $w$ & $w$ & data-free & - & generator + digestion\\
DaFKD \cite{wang2023dafkd} & NIID & $w$, $\theta^d$, $\theta^g$ & $w$, $\theta^g$ & data-free & generator & generator + digestion\\
FedGKD \cite{yao2021local}& NIID & $w$ & $w, w^{h}$ & data-free & regularizer  & - \\
FedNTD \cite{lee2021preservation} & NIID & $w$ & $w$ & data-free & regularizer  & - \\
FedLMD \cite{lu2023federated} & NIID & $w$ & $w$ & data-free & regularizer  & - \\
FedED \cite{guo2024not} & NIID & $w$ & $w$ & data-free & regularizer  & - \\
FedCAD \cite{he2022class} & NIID & $w$ & $w, \alpha_y$ & labeled & regularizer  & -\\
FedSSD \cite{he2022learning} & NIID & $w$ & $w, C$ & labeled & regularizer  & - \\
FedMLB \cite{kim2022multi} & NIID & $w$ & $w$ & data-free & regularizer  & - \\
FedBR\cite{liu2023adaptive} & NIID, CE & $w$ & $w$ & data-free & regularizer  & - \\
FedAlign \cite{mendieta2022local} & NIID & $w$ & $w$ & data-free & regularizer  & - \\
FedDistill$^{+}$ \cite{yao2021local} & NIID & $w , \overline{z}$ & $w , \overline{z}$ & data-free & regularizer & - \\ 
FedGen \cite{zhu2021data} & NIID &  $w, c$ & $w$ & data-free & generator & -\\
pFedSD \cite{jin2022personalized} & P & $w$ & $w$ & data-free & regularizer & - \\
\citeauthor{chen2023spectral} \cite{chen2023spectral} & P & $w$ & $w$ & data-free & regularizer & - \\
\citeauthor{wu2022federatedunlearning} \cite{wu2022federatedunlearning} & FU & $w$ & $w$ & unlabeled & - & digestion \\
FedET \cite{liu2023fedet} & CI, NIID & $w, c$ & $w$ & data-free & generator & generator + digestion \\
FedNASD \cite{wu2024federated} & CI, NIID & $w$ & $w$ & data-free & regularizer & - \\
\bottomrule
\end{tabular}
\caption{Synoptic overview of the surveyed solutions. We have identified 6 possible categories for primary purpose: model heterogeneity (MH), non-iidness (NIID), communication efficiency (CE), unlearning (FU), personalization (P), class-incremental learning (CI). Upload refers to the client-to-server link. Symbols: $w$ model parameters, 
$z$ logit vectors (model output before softmax), $\tilde{y}_p$ soft targets (model output after softmax) on public data,
$w^{h}$ historical model parameters,
$\overline{z}$ per-label average logit vectors,
$y$ labels of local data,
$H$ intermediate feature maps,
$A$ attention maps,
$\alpha_y$ per-class adaptive weights,
$C$ credibility matrix,
$c$ local label count,
$\theta^g$ generator model weights,
$\theta^d$ discriminator model weights.
We do not differentiate among model weights and model updates.}
\label{tab:large_comparison}
}
\end{center}
\end{table*}

\section{Enabling FL Model Heterogeneity via KD}
\label{sec:modelagnostic}
KD has been initially designed to transfer knowledge among neural networks with different structure and depth. In this Subsection, we review strategies that adopt KD to enable model agnosticism in FL, i.e., transfer knowledge among clients with heterogeneous model architectures. 

\subsection{Enhancing FedAvg Aggregation}
FedAvg's protocol can be enhanced to enable model heterogeneity by leveraging server-side ensemble distillation on top of the aggregation step \cite{lin2020ensemble,sattler2021fedaux}, through which knowledge is transferred among clients with different model architecture. To this end, the server can maintain a set of prototypical models, with each prototype representing all learners with same architecture. After collecting updates from clients, the server firstly performs a per-prototype model aggregation and then produces soft targets for each received client model either leveraging unlabeled data or synthetically generated examples. Next, such soft targets are averaged and used to fine tune each aggregated model prototype. 
Alternative possible solutions to enable model heterogeneity consist in exploiting distributed adaptations of CD instead of parameter-averaging algorithms such as FedAvg, as presented in the following. 

\subsection{Federated Adaptations of Codistillation}

\begin{figure}[t!]
\centering
\includegraphics[width=0.48\textwidth]{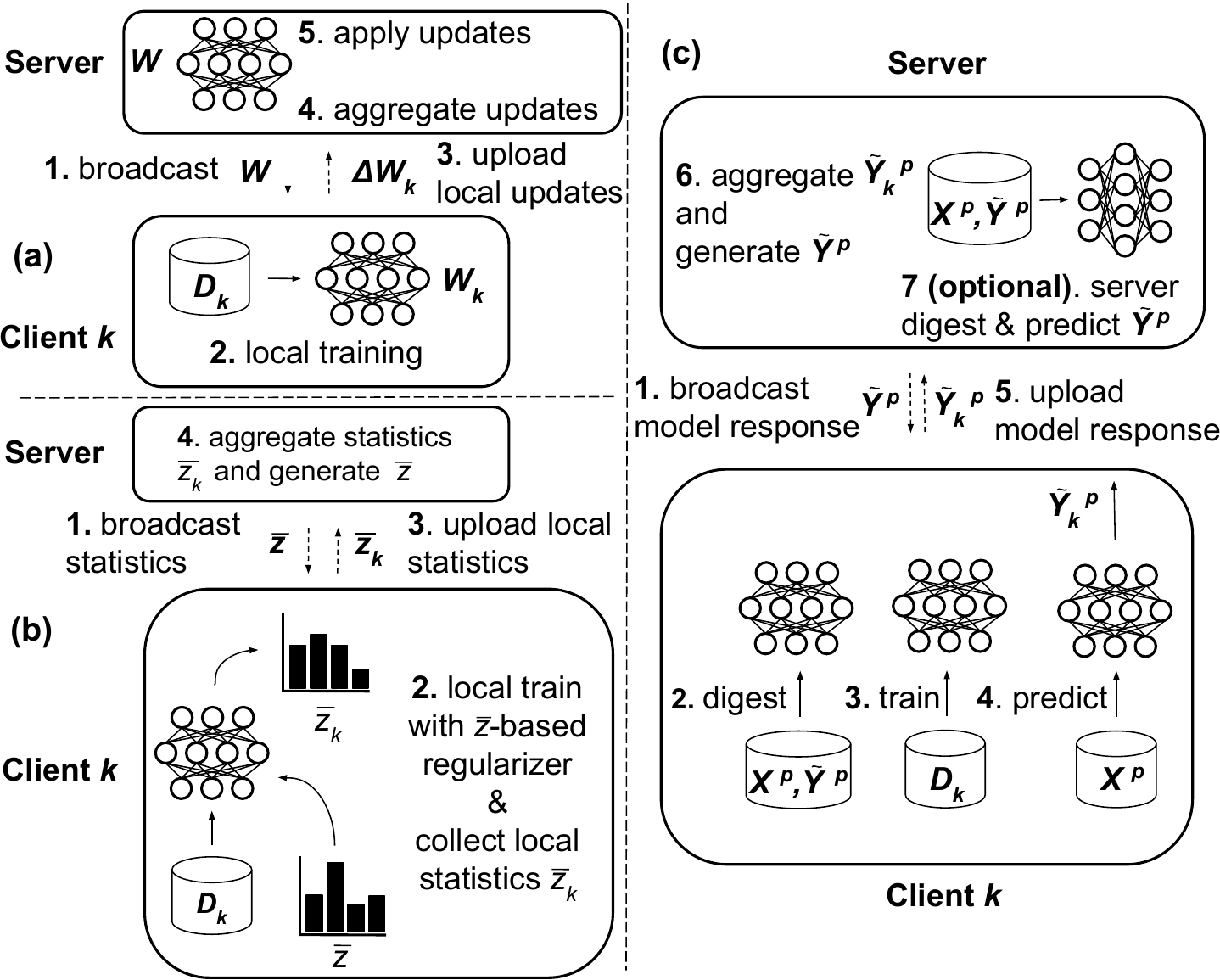}
\caption{KD-based solutions for FL issues. (a) FedAvg, (b) statistic-based federated CD, (c) response-based federated CD.}
\label{fig:comparison_favg_codistillation}
\end{figure}

In a general federated adaptation of CD \cite{anil2018large}, each client at round $t$ acts as student and sees the ensemble of clients knowledge at round $t - 1$ as a virtual teacher knowledge. As highlighted in Section \ref{subsec:codistillation}, traditional CD postulates that the worker networks have access to the same training dataset to form an ensemble of model responses on common samples. Such a requirement is not acceptable in the FL context, where collaborative training is performed without disclosing the private raw data of clients. Therefore, considering a classification task, federated adaptations of CD avoids the aforementioned problem of collecting model responses on common training data examples by exchanging a different kind of knowledge. Among the solutions in the literature, we identify three types of knowledge to enable federated versions of CD, and they are:
\begin{enumerate}
    \item ensemble of aggregated statistics of model responses on local data (e.g. per-label mean model responses),
    \item ensemble of local model responses computed on a publicly available dataset and not on local data,
    \item or ensemble of both model responses and model-agnostic intermediate features.
\end{enumerate}
It is worth noting that the clients and the server exchange this kind of information instead of model parameters. Furthermore, FL adaptations of CD relax the constraint of implementing the same model architecture, which was needed in regular CD for datacenter-oriented distributed training. In fact, exchanging knowledge based on model responses (or on model-agnostic intermediate features) enables heterogeneous models among workers, i.e., FL clients, as long as they have the same output shape. Fig. \ref{fig:comparison_favg_codistillation} illustrates FL adaptations of CD with respect to FedAvg baseline. Table \ref{tab:baseline_comparison} sums up the comparison among FL adaptations of CD treated in the following subsections.


\begin{table*}[h!]
\centering
{\footnotesize
\begin{tabular}{c c c c c}
\toprule
 & Knowledge & Transfer Set & Server Model & Notes \\ 
\bottomrule
FedDistill \cite{jeong2018communication} & statistics & data-free & no & KD-based regularizer \\
FedMD \cite{li2019fedmd} & response & labeled & no & Pre-training on $D_p$ \\
Cronus \cite{chang2019cronus} & response & unlabeled & no & Local training on $(X_p, \tilde{Y_p}) \cup D_k$ \\
DS-FL\cite{itahara2020distillation} & response & unlabeled & no & Entropy Reduction Aggregation \\
MATH \cite{hu2021mhat} & response & labeled & yes &  Server training on $(X_p, \tilde{Y_p}) \cup D_p$ \\
CFD \cite{sattler2021cfd} & response & unlabeled & yes & Compressed soft targets \\
FedGEMS \cite{cheng2021fedgems} & response & labeled & yes & Server-side kowledge refinement  \\
FedAD \cite{gong2021ensemble} & feature & unlabeled & yes & One-shot algorithm\\
FedGKT \cite{he2020group} & feature & data-free & yes & FL + Split learning paradigm\\
FedDKC \cite{wu2022exploring} & feature & data-free & yes & Knowledge refinement\\
\bottomrule
\end{tabular}
\caption{Comparison among strategies to enable model heterogeneity via FL adaptations of CD. $D_k$ represents the local private dataset of a generic client. $D_p$ represents a public transfer set. ($X_p$, $\tilde{Y}_p$) is the public transfer dataset labeled with soft targets $\tilde{Y}_p$. Knowledge column is inspired by the classification in [Guo \textit{et al.}, 2021]; statistics-based disclose aggregated statistics (e.g., per-label mean logit vector) of client model responses on local data, response-based methods communicate model outputs, feature-based also share intermediate representations.}
\label{tab:baseline_comparison}
}
\end{table*}

\paragraph{Disclosing Aggregated Statistics of Model Responses on Local Data.} In \cite{jeong2018communication}, Jeong et al. presented a pioneering distillation-based baseline for FL, FedDistill, depicted in Fig. \ref{fig:comparison_favg_codistillation}b. Participants periodically transmit only per-label mean soft targets computed on their private dataset. The server, in turn, averages such tensors and produces per-label global soft targets to be broadcast the next round. When locally training, clients regularize their local loss with a per-label distillation term which uses the received global soft targets as the teacher's output. A similar strategy is later presented in \cite{seo2020federated}. It is worth noting that FedDistill is extremely communication efficient with respect to parameter-based schemes when considering DL models for classification task, since the communication payload size depends on the model response size and not on model size.

\paragraph{Exchanging Model Responses on Publicly Available Data.} Federated CD can be enabled by using knowledge formed by an ensemble of model responses computed on a proxy transfer set (publicly available, both clients and server can retrieve it). In this way, clients train on their private data, and share knowledge via their model response on the transfer set. Here, the approaches in literature are more variegated, but a general skeleton of algorithmic steps can be the following (illustrated in Fig. \ref{fig:comparison_favg_codistillation}c): 
\begin{enumerate}
    \item \textit{broadcast}: clients receive the current global logits/soft targets;
    \item \textit{local digest}: clients distill their local model by mimicking the received global logits/soft-labels on a subset of the transfer dataset. Here, the parallel with traditional CD: each client sees the averaged predictions of the other clients at the previous round as a virtual teacher. This step can be also seen as a way to retrieve the global model parameters instead of directly receiving them from the server as in parameter-based schemes (e.g., FedAvg);
    \item \textit{revisit (local train)}: clients fine-tune the distilled model on local data;
    \item \textit{local predict}: clients compute their local logits/soft targets on a subset of the transfer dataset;
    \item \textit{upload}: clients sends back the computed logits/soft targets;
    \item \textit{aggregate}: the server aggregates the client predictions to produce the updated global logits/soft targets. 
    \item \textit{(optional) global digest}: the server distills a model from the aggregated soft targets on the respective subset of the transfer dataset, and uses it again to generate the global logits/soft targets to broadcast. Learning a server-side model, that is refined round by round, can improve the training process when there is partial participation of clients. Next, a new round begins.
\end{enumerate}
Table \ref{tab:baseline_comparison} summarizes the deviation of the surveyed works from the above general algorithmic steps. In particular, considering either a labeled or unlabeled proxy dataset influences the design of algorithms. FedMD \cite{li2019fedmd} uses a proxy labeled dataset to perform an initial pretraining phase on clients, before the protocol starts.
\citeauthor{itahara2020distillation} modify the aggregation step proposing an Entropy Reduction Aggregation (ERA), demonstrating that using a temperature lower than 1 when applying softmax to the aggregated logits reduces the entropy of global soft targets, and can help the training process, especially in non-IID settings \cite{itahara2020distillation}.
Compressed Federated Distillation (CFD) \cite{sattler2021cfd} implements an extreme and effective compression technique for soft targets based on quantization and delta coding, which is applied both by clients and server before communicating. 
Cronus \cite{chang2019cronus} merges the \textit{local digest} and \textit{revisit} step by directly training on the union (i.e., concatenation) of the private dataset and the soft-labeled public one. In addition, Cronus aggregates soft targets following the approach in \cite{diakonikolas2017being} for enhanced robustness. Similarly to Cronus, in MATH \cite{hu2021mhat} clients jointly train on private dataset, public dataset, and public dataset tagged with global soft targets. MATH \cite{hu2021mhat} considers a labeled proxy dataset, and distills its server model by training it on the union of such a public dataset with the soft-labeled version of it.
FedGEM \cite{cheng2021fedgems} proposes to employ a powerful model server, and its variation FedGEMS exploits the labels in the public transfer set to enforce a selection and weighting strategy which can improve the server-side digestion \cite{cheng2021fedgems}.

\paragraph{Leveraging Intermediate Features.} 
FedAD \cite{gong2021ensemble} also uses intermediate features besides model output to extend response-based knowledge distillation. The intermediate features are model-agnostic attention maps \cite{selvaraju2017grad}, which still enable model heterogeneity as long as there is consensus on attention map shape. FedAD is a one-shot federated learning framework, which means that clients do not have to distill their local model at the beginning of each round, and can participate asynchronously. FedGKT \cite{he2020group} uses intermediate features tacking advantage of both asynchronous split learning paradigm \cite{poirot2019split} and regular FL.
Edge devices train small networks composed of a feature extractor, which produces intermediate feature maps, and a classifier, which produces soft targets. Similarly, the server leverages a deeper network and a classifier. After local training, for each local samples, clients communicate their computed intermediate features, the predicted soft targets and the related ground truth labels. The server takes locally computed extracted features as input for its deeper network and produces global soft targets. Both clients and server use a linear combination of regular cross-entropy loss and KD-based loss as objective function. The first considers soft targets and ground truth labels, the latter measures the discrepancy among local and global logits. A similar framework is implemented and extended in FedDKC \cite{wu2022exploring}, where \citeauthor{wu2022exploring} also develop server-side knowledge refinement strategies. 

\subsection{Comparison and Adoption Guidelines} Federated adaptations of CD can enable model heterogeneity, and can reduce the communication requirements at the cost of computation overhead with respect to parameter-based schemes. Hence, despite being extremely communication efficient, it may be not always possible to deploy such algorithms on resource-constrained devices due to the overhead of client-side distillation  (in Table \ref{tab:large_comparison}, solutions which use \textit{digestion} at client side), while being a suitable model-agnostic alternative for cross-silo settings.\footnote{For example, \citeauthor{sattler2021cfd} use 80000 data points from a public dataset to distill on-device model before local training.} Furthermore, this class of solutions usually performs worse than FedAvg-based baselines (in terms of global model accuracy) \cite{sattler2021fedaux} -- even though they typically improve the performance of non-collaborative training \cite{itahara2020distillation}. Moreover, most works in this category suppose the existence of a semantically-similar proxy dataset (in some cases even labeled), which may be an unrealistic assumption in some deployment scenarios and use cases (e.g., for specific medical applications). The pioneering communication-efficient data-free strategy in \cite{jeong2018communication} does not incur in local computation overhead, but it is far from achieving global model test accuracy comparable to FedAvg, as demonstrated in \cite{zhu2021data}, also disclosing possible privacy-sensitive information about private data (i.e., per-label model outputs). Solutions as \cite{he2020group,wu2022exploring} enable model heterogeneity, are usually more communication efficient than FedAvg, and include resource-constrained devices in the federation, by adopting a split-learning paradigm and by taking advantage of KD-based regularization. However, as shown in Table \ref{tab:large_comparison}, due to their split-learning approach, these solutions disclose local ground-truth labels, which again may incur in privacy violation.

\section{Tackling FL Data Heterogeneity via KD}
\label{sec:dataagnostic}
KD-based solutions can be used to improve the generalization performance of the global model in presence of data heterogeneity. Server-side approaches rectify FedAvg's global model either via ensemble distillation on a proxy dataset or using a data-free generator. Orthogonally. client-side mechanisms limit local overfitting or directly control the phenomenon of client drift by distilling global knowledge via on-device regularizers or synthetically-generated data.
\subsection{Server-side Refinement of Global Model}
\label{sec:serverside_refinement}
\paragraph{Refinement on Pubicly Available Data.}
In \cite{lin2020ensemble}, the authors propose FedDF, a server-side ensemble distillation approach to both enable model heterogeneity and enhance FedAvg's aggregation. In FedDF, the global model is fine tuned imitating the ensemble (e.g., weighted average) of clients' model output on a proxy dataset. FedAUX \cite{sattler2021fedaux} boosts the performances of FedDF \cite{lin2020ensemble} leveraging unsupervised pre-training on auxiliary data to find a suitable model initialization for client-side feature extractor. In addition, FedAUX weights the ensemble predictions on the proxy data according to $(\epsilon, \delta)$-differentially private \cite{dwork2014algorithmic} certainty score of each participant model. FedBE \cite{chen2020fedbe} proposes to combine client predictions by means of a Bayesian model ensemble to further improve robustness of the aggregation.

\paragraph{Data-free Generator.} While server-side ensemble distillation approaches  suppose the existence of a proxy dataset, FedFTG \cite{zhang2022fine} performs a server-side refinement of the global model via data-free knowledge distillation where the server adversarially trains both a generator model and the global model, and fine-tunes the latter with synthetic data. A data-free generator-based refinement of global model is also proposed in \cite{zhang2021fedzkt}. The recently proposed DaFKD \cite{wang2023dafkd} leverages both a server-side and a client-side generator. At each FL round, the server refines the FedAvg-aggregated global model using an importance-weighted ensemble of client models as teacher. The refined global model is distilled with a loss of the form:  
\begin{equation}
\label{eq:dafkd}
    \mathcal{L} =  KL(\sum_{k}^{K}\alpha_k \cdot \hat{y}^{w
    ^{k}}, \hat{y}^w),
\end{equation}
with the activated $K$ clients indexed by $k$, and $\hat{y}^{w^{k}}$ and $\hat{y}^w$ being, respectively, the prediction of client $k$'s model and of the global model on synthetically generated data. $\alpha_k$ weights per-sample importance of client models' predictions according to a discriminator model, which is adversarially trained by each client to distinguish whether the synthetic data is sampled from the distribution of its local data.


\subsection{Client-side regularization}
\label{subsec:localglobaldistill}
\paragraph{Local Regularization to Reduce Overfitting.} In \cite{mendieta2022local}, \citeauthor{mendieta2022local} show that GradAug \cite{yang2020gradaug}, a distillation-based structural regularization not specifically designed for FL settings, effectively mitigates client drift issues though substantially introducing computation overhead. Hence, \citeauthor{mendieta2022local} design a novel method, FedAlign \cite{mendieta2022local}, which has similar effect and performances but with a sustainable computation overhead. In particular, FedAlign targets the deeper layers of a neural network, most prone to overfit client distribution \cite{luo2021no}, imposing a distillation-based term in the local objective function. 
Such term minimizes the discrepancy among the intermediate features produced in output by the final block of the full network and the features produced by the same block but at reduced width (via temporary uniform pruning). 
Employing a slimmed sub-block permits to introduce a limited computation overhead.

\begin{figure}
\centering
\includegraphics[width=0.32\textwidth]{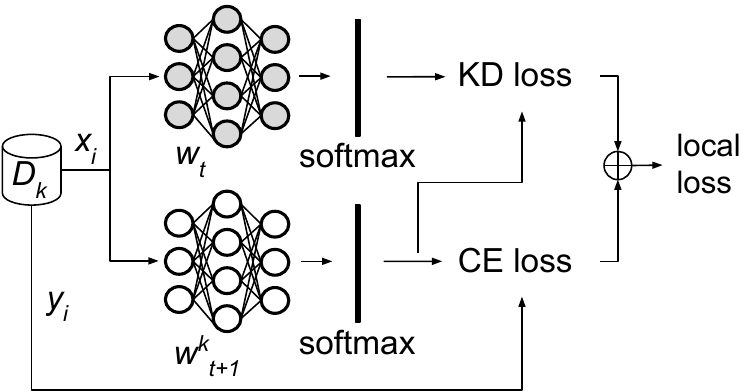}
\caption{Local-global distillation using a regularization term. $w_{t}$ represents the global model at round $t$. $w_{t+1}^k$ is the local model. }
\label{fig:distill_regularizer}
\end{figure}

\paragraph{Local-Global Distillation via Regularization Term.} 
Respectively inspired by fine-tuning optimization ideas and continual learning research, the recent works in \cite{yao2021local,ni2022federated} and \cite{lee2021preservation} find that local KD-based regularization is an effective way of reducing the influence of non-IID data in FL settings. 
In local-global distillation, the local objective function of clients becomes a linear combination between the cross-entropy loss and a KD-based loss,
\begin{equation}
\label{eq:localglobal}
    \mathcal{L} =  (1- \lambda) \mathcal{L}_{CE}(\tilde{y}^{w_{t+1}^{k}}, y) + \lambda \tau^2\mathcal{L}_{KD}(\tilde{y}^{w_{t+1}^{k}}, \tilde{y}^{w_{t}})
\end{equation}

where $\tilde{y}^{w_{t+1}^{k}}$ and $\tilde{y}^{w_{t}}$ are the soft targets produced on local data respectively by the local model of client $k$ and by the received global model.
The KD-based loss measures the discrepancy among the global model's output (i.e., the teacher model's output) and the local model's output (i.e., the student model output) on private data, e.g. via KL divergence, and works as a regularization term. In this way, the global model works as an anchor for local training, limiting client drift. Local-global distillation can be seen as a self-distillation mechanism, since the local and global model coincide at the beginning of local training. Fig. \ref{fig:distill_regularizer} depicts the basic framework for this kind of local-global distillation. 

\paragraph{Local-Global Distillation via Regularization Term: Further Improvements.} FedGKD \cite{yao2021local} uses an ensemble of $M$ historical global models as teacher, computed as the average of $M$ past global models. \citeauthor{yao2021local} also propose FedGKD-VOTE as a variation that considers the averaged discrepancy of all the $M$ historical models' outputs as the regularization term. With $M = 1$, the communication cost is the same of FedAvg, while for $M > 1$ the server-client communication cost is doubled, and for FedGKD-VOTE it scales with $M$. To reduce forgetting among subsequent rounds of learning, FedNTD \cite{lee2021preservation} ignores the logits produced by the true classes when computing the softmax score later fed to the KD-based loss. FedLMD \cite{lu2023federated} extends FedNTD by masking out the locally most represented classes in the teacher's logits. \citeauthor{lu2023federated} also proposes a teacher-free version, uniformly distributing the teacher's output probability on minority classes. Similarly, \citeauthor{guo2024not} design FedED that preserves the global knowledge on not locally represented classes (classes for which a client does not hold samples) \cite{guo2024not}. \citeauthor{he2022class} further observe that, in the framework of Fig. \ref{fig:distill_regularizer}, leveraging an inaccurate global model (i.e., inaccurate teacher) on specific classification classes might mislead local training \cite{he2022class}. To alleviate such phenomenon, a class-wise adaptive weight is proposed in FedCAD \cite{he2022class} to control the impact of distillation: when the global model is accurate on a certain class, local models learn more from the distilled knowledge. FedCAD determines the class-wise adaptive weight based on the performances of the global model on an auxiliary dataset, and the server broadcasts such information along with model parameters round by round. FedSSD \cite{he2022learning} extends FedCAD by also considering the credibility of global model at the instance level when computing the distillation term in local training. FedMLB \cite{kim2022multi} enhances the local-global distillation also using intermediate representations, preventing them from deviating too significantly during local fine tuning. To this end, FedMLB crafts hybrid pathways composed of local and global subnetworks, i.e., of local network blocks followed by non-trainable global blocks. Besides regular cross-entropy, local learning also considers the average cross-entropy from hybrid paths and the average KL divergence between the outputs produced by the hybrid paths and the main path as regularization term. FedBR, proposed in \cite{liu2023adaptive}, extends FedMLB by adaptively distributing only a fraction of the global model's blocks according to the estimated level of data heterogeneity and on the client's computation/communication constraints. FedDistill$^{+}$, used as alternative baseline in \cite{yao2021local,zhu2021data}, extends the work of \cite{jeong2018communication,seo2020federated} by exchanging per-label local logits on training dataset in addition to model parameters. With respect to the framework in Fig. \ref{fig:distill_regularizer}, FedDistill$^{+}$ uses the received per-label globally averaged logits instead of the output of the global model on private data to calculate the KD loss.
\paragraph{Local-global Distillation via Data-Free Generator Models.}  Differently from the other work in this subsection, FedGen \cite{zhu2021data} learns a lightweight server-side generator which is distributed, round by round, to clients that sample it to obtain augmented training examples, using global knowledge as inductive biases in local learning. 

\subsection{Comparison and Adoption Guidelines}  
KD-based server-side refinement strategies such as \cite{lin2020ensemble,sattler2021fedaux,chen2020fedbe} can improve FedAvg global model performance in presence of highly heterogeneous data when semantically-similar unlabeled proxy data are available. It is worth noting that this class of algorithms exhibits most improvements when several local epochs are performed between communication rounds and client models tend to drift apart. Also, such algorithms do not introduce computation or communication overhead on clients. Data-free generator models can also be used to perform server-side global model corrections as in \cite{zhang2022fine} or to limit client drift directly at the participating devices as in \cite{zhu2021data}, in both cases at the cost of disclosing local label count. Similarly, DaFKD \cite{wang2023dafkd} enables data-free refinement of global model via generators, but introducing computation and communication overheads on clients -- although limited thanks to partial parameter sharing.

No additional information has to be disclosed from clients and not even proxy data are needed in solutions that regularize local training by employing global model output on local data, as in FedGKD \cite{yao2021local}, FedNTD \cite{lee2021preservation}, FedLMD \cite{lu2023federated}, FedED \cite{guo2024not}. In addition, this set of strategies do not introduce significant on-device computation overhead and has the same communication requirements as FedAvg. However, they would require to store two full-size models in memory (the local model as usual and the global model as reference), such limitation can, in practice, be avoided by firstly computing the predictions of the received global model on local data, and then proceeding with local training by overwriting it. If limited labeled proxy data are available, local-global knowledge distillation can be improved as in \cite{he2022class,he2022learning}. When moderate or even significant computing overhead is sustainable, local global distillation can be enhanced by using intermediate features and hybrid pathways as in FedMLB \cite{kim2022multi} and FedBR \cite{liu2023adaptive}, thus significantly improving the effectiveness of local-global distillation. It is worth noting that in FedMLB and FedBR the selected blocks of the global model must be stored locally during training, and this cannot be avoided as for regularization based on model responses. As highlighted in \cite{kim2022multi}, client-side regularization can be coupled with standard server-side strategies to boost performances (e.g., FedOpt \cite{reddi2020adaptive}). 

\section{Emerging Use of KD in FL}
\label{sec:future_dir}
Here we outline the recently emerging use of KD tailored to address other FL issues, i.e., personalized FL, federated unlearning, and class-incremental federated learning.

\paragraph{KD for FL Personalization.}
The primary scope of personalized FL is to build a global model prone to quickly adapt to local data distributions. While KD-based solutions are commonly used to enhance the generalization ability of the global model (see Section \ref{sec:dataagnostic}), lately, distillation-inspired mechanisms have been employed in personalized FL algorithms. \citeauthor{jin2022personalized} propose pFedSD where clients store their last trained personalized model and use it as teacher for the next rounds \cite{jin2022personalized}. In a nutshell, pFedSD uses the distillation framework depicted in Fig. \ref{fig:distill_regularizer}, with the teacher being a past local model instead of the global model. 
In addition, \citeauthor{chen2023spectral} recently introduced a \textit{spectral} distillation used to regularize local training via a term that minimizes the KL divergence between the Fourier spectra of global and personalized models \cite{chen2023spectral}. 

\paragraph{KD for Federated Unlearning.} An FL client should have the right to request the removal of its contribution from the global model, and mechanisms to enable selective forgetting of clients are emerging. The solution in \cite{wu2022federatedunlearning} firstly removes the unlearning client's historical model updates from the global model. Then, the server leverages KD to quickly recover the global model performance with the sanitized global model mimicking the outputs of the last global model on proxy unlabeled data. 

\paragraph{KD for Federated Class-Incremental Learning.} The methods surveyed up to here assume that the classification classes of a task are fixed over the entire FL process. In contrast, works to address class-incremental learning in federated settings are emerging. In \citeauthor{liu2023fedet}, \cite{liu2023fedet} propose FedET, which leverages an enhancer distillation method to modify the imbalance between old and new knowledge. In \cite{wu2024federated}, \citeauthor{wu2024federated} design FedNASD, a method that uses the class probabilities from the current models to approximate the historical models' output on new classes -- historical models may encompass fewer classes -- so that local-global distillation (see Fig. \ref{fig:distill_regularizer}) can be applied even when classes are changing over time.

\section{Conclusive Remarks}

KD has been recently embedded in FL algorithms to tackle FL-specific issues. In this survey, we present the various possible distillation-based mechanisms primarily proposed to enable model heterogeneity and to tackle the effects of data heterogeneity in FL, and, in some cases, to reduce the associated communication costs. We originally classify and compare the recent proposals in this field, by highlighting their possible weaknesses and their trade-offs. Then, we outline the emerging use of KD to improve other aspects of FL.  

\appendix





\newpage
\bibliographystyle{named}
\bibliography{ijcai24}

\begin{thebibliography}{}

\bibitem[\protect\citeauthoryear{Anil \bgroup \em et al.\egroup }{2018}]{anil2018large}
Rohan Anil, Gabriel Pereyra, Alexandre Passos, Robert Ormandi, George~E Dahl, and Geoffrey~E Hinton.
\newblock Large scale distributed neural network training through online distillation.
\newblock {\em arXiv preprint arXiv:1804.03235}, 2018.

\bibitem[\protect\citeauthoryear{Bellavista \bgroup \em et al.\egroup }{2021}]{bellavista2021decentralised}
Paolo Bellavista, Luca Foschini, and Alessio Mora.
\newblock Decentralised learning in federated deployment environments: A system-level survey.
\newblock {\em ACM Computing Surveys (CSUR)}, 54(1):1--38, 2021.

\bibitem[\protect\citeauthoryear{Chang \bgroup \em et al.\egroup }{2019}]{chang2019cronus}
Hongyan Chang, Virat Shejwalkar, Reza Shokri, and Amir Houmansadr.
\newblock Cronus: Robust and heterogeneous collaborative learning with black-box knowledge transfer.
\newblock {\em arXiv preprint arXiv:1912.11279}, 2019.

\bibitem[\protect\citeauthoryear{Chen and Chao}{2020}]{chen2020fedbe}
Hong-You Chen and Wei-Lun Chao.
\newblock Fedbe: Making bayesian model ensemble applicable to federated learning.
\newblock {\em arXiv preprint arXiv:2009.01974}, 2020.

\bibitem[\protect\citeauthoryear{Chen \bgroup \em et al.\egroup }{2023}]{chen2023spectral}
Zihan Chen, Howard~Hao Yang, Tony Quek, and K.~F.~Ernest Chong.
\newblock Spectral co-distillation for personalized federated learning.
\newblock In {\em Thirty-seventh Conference on Neural Information Processing Systems}, 2023.

\bibitem[\protect\citeauthoryear{Cheng \bgroup \em et al.\egroup }{2021}]{cheng2021fedgems}
Sijie Cheng, Jingwen Wu, Yanghua Xiao, and Yang Liu.
\newblock Fedgems: Federated learning of larger server models via selective knowledge fusion.
\newblock {\em arXiv preprint arXiv:2110.11027}, 2021.

\bibitem[\protect\citeauthoryear{Diakonikolas \bgroup \em et al.\egroup }{2017}]{diakonikolas2017being}
Ilias Diakonikolas, Gautam Kamath, Daniel~M Kane, Jerry Li, Ankur Moitra, and Alistair Stewart.
\newblock Being robust (in high dimensions) can be practical.
\newblock In {\em International Conference on Machine Learning}, pages 999--1008. PMLR, 2017.

\bibitem[\protect\citeauthoryear{Dwork \bgroup \em et al.\egroup }{2014}]{dwork2014algorithmic}
Cynthia Dwork, Aaron Roth, et~al.
\newblock The algorithmic foundations of differential privacy.
\newblock {\em Foundations and Trends{\textregistered} in Theoretical Computer Science}, 9(3--4):211--407, 2014.

\bibitem[\protect\citeauthoryear{Gong \bgroup \em et al.\egroup }{2021}]{gong2021ensemble}
Xuan Gong, Abhishek Sharma, Srikrishna Karanam, Ziyan Wu, Terrence Chen, David Doermann, and Arun Innanje.
\newblock Ensemble attention distillation for privacy-preserving federated learning.
\newblock In {\em Proceedings of the IEEE/CVF International Conference on Computer Vision}, pages 15076--15086, 2021.

\bibitem[\protect\citeauthoryear{Gou \bgroup \em et al.\egroup }{2021}]{gou2021knowledge}
Jianping Gou, Baosheng Yu, Stephen~J Maybank, and Dacheng Tao.
\newblock Knowledge distillation: A survey.
\newblock {\em International Journal of Computer Vision}, 129(6):1789--1819, 2021.

\bibitem[\protect\citeauthoryear{Guo \bgroup \em et al.\egroup }{2024}]{guo2024not}
Kuangpu Guo, Yuhe Ding, Jian Liang, Ran He, Zilei Wang, and Tieniu Tan.
\newblock Not all minorities are equal: Empty-class-aware distillation for heterogeneous federated learning.
\newblock {\em arXiv preprint arXiv:2401.02329}, 2024.

\bibitem[\protect\citeauthoryear{He \bgroup \em et al.\egroup }{2020}]{he2020group}
Chaoyang He, Murali Annavaram, and Salman Avestimehr.
\newblock Group knowledge transfer: Federated learning of large cnns at the edge.
\newblock {\em Advances in Neural Information Processing Systems}, 33:14068--14080, 2020.

\bibitem[\protect\citeauthoryear{He \bgroup \em et al.\egroup }{2022a}]{he2022learning}
Yuting He, Yiqiang Chen, XiaoDong Yang, Hanchao Yu, Yi-Hua Huang, and Yang Gu.
\newblock Learning critically: Selective self-distillation in federated learning on non-iid data.
\newblock {\em IEEE Transactions on Big Data}, 2022.

\bibitem[\protect\citeauthoryear{He \bgroup \em et al.\egroup }{2022b}]{he2022class}
Yuting He, Yiqiang Chen, Xiaodong Yang, Yingwei Zhang, and Bixiao Zeng.
\newblock Class-wise adaptive self distillation for heterogeneous federated learning.
\newblock 2022.

\bibitem[\protect\citeauthoryear{Hinton \bgroup \em et al.\egroup }{2015}]{hinton2015distilling}
Geoffrey Hinton, Oriol Vinyals, and Jeff Dean.
\newblock Distilling the knowledge in a neural network.
\newblock {\em arXiv preprint arXiv:1503.02531}, 2015.

\bibitem[\protect\citeauthoryear{Hu \bgroup \em et al.\egroup }{2021}]{hu2021mhat}
Li~Hu, Hongyang Yan, Lang Li, Zijie Pan, Xiaozhang Liu, and Zulong Zhang.
\newblock Mhat: an efficient model-heterogenous aggregation training scheme for federated learning.
\newblock {\em Information Sciences}, 560:493--503, 2021.

\bibitem[\protect\citeauthoryear{Itahara \bgroup \em et al.\egroup }{2021}]{itahara2020distillation}
Sohei Itahara, Takayuki Nishio, Yusuke Koda, Masahiro Morikura, and Koji Yamamoto.
\newblock Distillation-based semi-supervised federated learning for communication-efficient collaborative training with non-iid private data.
\newblock {\em IEEE Transactions on Mobile Computing}, 22(1):191--205, 2021.

\bibitem[\protect\citeauthoryear{Jeong \bgroup \em et al.\egroup }{2018}]{jeong2018communication}
Eunjeong Jeong, Seungeun Oh, Hyesung Kim, Jihong Park, Mehdi Bennis, and Seong-Lyun Kim.
\newblock Communication-efficient on-device machine learning: Federated distillation and augmentation under non-iid private data.
\newblock {\em arXiv preprint arXiv:1811.11479}, 2018.

\bibitem[\protect\citeauthoryear{Jin \bgroup \em et al.\egroup }{2022}]{jin2022personalized}
Hai Jin, Dongshan Bai, Dezhong Yao, Yutong Dai, Lin Gu, Chen Yu, and Lichao Sun.
\newblock Personalized edge intelligence via federated self-knowledge distillation.
\newblock {\em IEEE Transactions on Parallel and Distributed Systems}, 34(2):567--580, 2022.

\bibitem[\protect\citeauthoryear{Karimireddy \bgroup \em et al.\egroup }{2020}]{karimireddy2019scaffold}
Sai~P Karimireddy, Satyen Kale, Mehryar Mohri, Sashank Reddi, Sebastian Stich, and A~Theertha Suresh.
\newblock Scaffold: Stochastic controlled averaging for federated learning.
\newblock In {\em International Conference on Machine Learning}, pages 5132--5143. PMLR, 2020.

\bibitem[\protect\citeauthoryear{Kim \bgroup \em et al.\egroup }{2022}]{kim2022multi}
Jinkyu Kim, Geeho Kim, and Bohyung Han.
\newblock Multi-level branched regularization for federated learning.
\newblock In {\em International Conference on Machine Learning}, pages 11058--11073. PMLR, 2022.

\bibitem[\protect\citeauthoryear{Lee \bgroup \em et al.\egroup }{2022}]{lee2021preservation}
Gihun Lee, Minchan Jeong, Yongjin Shin, Sangmin Bae, and Se-Young Yun.
\newblock Preservation of the global knowledge by not-true distillation in federated learning.
\newblock In {\em Advances in Neural Information Processing Systems}, 2022.

\bibitem[\protect\citeauthoryear{Li and Wang}{2019}]{li2019fedmd}
Daliang Li and Junpu Wang.
\newblock Fedmd: Heterogenous federated learning via model distillation.
\newblock {\em arXiv preprint arXiv:1910.03581}, 2019.

\bibitem[\protect\citeauthoryear{Lin \bgroup \em et al.\egroup }{2020}]{lin2020ensemble}
Tao Lin, Lingjing Kong, Sebastian~U Stich, and Martin Jaggi.
\newblock Ensemble distillation for robust model fusion in federated learning.
\newblock {\em Advances in Neural Information Processing Systems}, 33:2351--2363, 2020.

\bibitem[\protect\citeauthoryear{Liu \bgroup \em et al.\egroup }{2023a}]{liu2023fedet}
Chenghao Liu, Xiaoyang Qu, Jianzong Wang, and Jing Xiao.
\newblock Fedet: a communication-efficient federated class-incremental learning framework based on enhanced transformer.
\newblock In {\em Proceedings of the Thirty-Second International Joint Conference on Artificial Intelligence}, pages 3984--3992, 2023.

\bibitem[\protect\citeauthoryear{Liu \bgroup \em et al.\egroup }{2023b}]{liu2023adaptive}
Jianchun Liu, Qingmin Zeng, Hongli Xu, Yang Xu, Zhiyuan Wang, and He~Huang.
\newblock Adaptive block-wise regularization and knowledge distillation for enhancing federated learning.
\newblock {\em IEEE/ACM Transactions on Networking}, 2023.

\bibitem[\protect\citeauthoryear{Lu \bgroup \em et al.\egroup }{2023}]{lu2023federated}
Jianghu Lu, Shikun Li, Kexin Bao, Pengju Wang, Zhenxing Qian, and Shiming Ge.
\newblock Federated learning with label-masking distillation.
\newblock In {\em Proceedings of the 31st ACM International Conference on Multimedia}, pages 222--232, 2023.

\bibitem[\protect\citeauthoryear{Luo \bgroup \em et al.\egroup }{2021}]{luo2021no}
Mi~Luo, Fei Chen, Dapeng Hu, Yifan Zhang, Jian Liang, and Jiashi Feng.
\newblock No fear of heterogeneity: Classifier calibration for federated learning with non-iid data.
\newblock {\em Advances in Neural Information Processing Systems}, 34:5972--5984, 2021.

\bibitem[\protect\citeauthoryear{McMahan \bgroup \em et al.\egroup }{2016}]{mcmahan2016communication}
H~Brendan McMahan, Eider Moore, Daniel Ramage, Seth Hampson, et~al.
\newblock Communication-efficient learning of deep networks from decentralized data.
\newblock {\em arXiv preprint arXiv:1602.05629}, 2016.

\bibitem[\protect\citeauthoryear{Mendieta \bgroup \em et al.\egroup }{2022}]{mendieta2022local}
Matias Mendieta, Taojiannan Yang, Pu~Wang, Minwoo Lee, Zhengming Ding, and Chen Chen.
\newblock Local learning matters: Rethinking data heterogeneity in federated learning.
\newblock In {\em Proceedings of the IEEE/CVF Conference on Computer Vision and Pattern Recognition}, pages 8397--8406, 2022.

\bibitem[\protect\citeauthoryear{Ni \bgroup \em et al.\egroup }{2022}]{ni2022federated}
Xuanming Ni, Xinyuan Shen, and Huimin Zhao.
\newblock Federated optimization via knowledge codistillation.
\newblock {\em Expert Systems with Applications}, 191:116310, 2022.

\bibitem[\protect\citeauthoryear{Papernot \bgroup \em et al.\egroup }{2016}]{papernot2016semi}
Nicolas Papernot, Mart{\'\i}n Abadi, Ulfar Erlingsson, Ian Goodfellow, and Kunal Talwar.
\newblock Semi-supervised knowledge transfer for deep learning from private training data.
\newblock {\em arXiv preprint arXiv:1610.05755}, 2016.

\bibitem[\protect\citeauthoryear{Poirot \bgroup \em et al.\egroup }{2019}]{poirot2019split}
Maarten~G Poirot, Praneeth Vepakomma, Ken Chang, Jayashree Kalpathy-Cramer, Rajiv Gupta, and Ramesh Raskar.
\newblock Split learning for collaborative deep learning in healthcare.
\newblock {\em arXiv preprint arXiv:1912.12115}, 2019.

\bibitem[\protect\citeauthoryear{Reddi \bgroup \em et al.\egroup }{2020}]{reddi2020adaptive}
Sashank Reddi, Zachary Charles, Manzil Zaheer, Zachary Garrett, Keith Rush, Jakub Kone{\v{c}}n{\`y}, Sanjiv Kumar, and H~Brendan McMahan.
\newblock Adaptive federated optimization.
\newblock {\em arXiv preprint arXiv:2003.00295}, 2020.

\bibitem[\protect\citeauthoryear{Sattler \bgroup \em et al.\egroup }{2019}]{sattler2019robust}
Felix Sattler, Simon Wiedemann, Klaus-Robert M{\"u}ller, and Wojciech Samek.
\newblock Robust and communication-efficient federated learning from non-iid data.
\newblock {\em IEEE transactions on neural networks and learning systems}, 2019.

\bibitem[\protect\citeauthoryear{Sattler \bgroup \em et al.\egroup }{2021a}]{sattler2021fedaux}
Felix Sattler, Tim Korjakow, Roman Rischke, and Wojciech Samek.
\newblock Fedaux: Leveraging unlabeled auxiliary data in federated learning.
\newblock {\em IEEE Transactions on Neural Networks and Learning Systems}, 2021.

\bibitem[\protect\citeauthoryear{Sattler \bgroup \em et al.\egroup }{2021b}]{sattler2021cfd}
Felix Sattler, Arturo Marban, Roman Rischke, and Wojciech Samek.
\newblock Cfd: Communication-efficient federated distillation via soft-label quantization and delta coding.
\newblock {\em IEEE Transactions on Network Science and Engineering}, 2021.

\bibitem[\protect\citeauthoryear{Selvaraju \bgroup \em et al.\egroup }{2017}]{selvaraju2017grad}
Ramprasaath~R Selvaraju, Michael Cogswell, Abhishek Das, Ramakrishna Vedantam, Devi Parikh, and Dhruv Batra.
\newblock Grad-cam: Visual explanations from deep networks via gradient-based localization.
\newblock In {\em Proceedings of the IEEE international conference on computer vision}, pages 618--626, 2017.

\bibitem[\protect\citeauthoryear{Seo \bgroup \em et al.\egroup }{2020}]{seo2020federated}
Hyowoon Seo, Jihong Park, Seungeun Oh, Mehdi Bennis, and Seong-Lyun Kim.
\newblock Federated knowledge distillation.
\newblock {\em arXiv preprint arXiv:2011.02367}, 2020.

\bibitem[\protect\citeauthoryear{Wang \bgroup \em et al.\egroup }{2023}]{wang2023dafkd}
Haozhao Wang, Yichen Li, Wenchao Xu, Ruixuan Li, Yufeng Zhan, and Zhigang Zeng.
\newblock Dafkd: Domain-aware federated knowledge distillation.
\newblock In {\em Proceedings of the IEEE/CVF Conference on Computer Vision and Pattern Recognition}, pages 20412--20421, 2023.

\bibitem[\protect\citeauthoryear{Wu \bgroup \em et al.\egroup }{2022a}]{wu2022federatedunlearning}
Chen Wu, Sencun Zhu, and Prasenjit Mitra.
\newblock Federated unlearning with knowledge distillation.
\newblock {\em arXiv preprint arXiv:2201.09441}, 2022.

\bibitem[\protect\citeauthoryear{Wu \bgroup \em et al.\egroup }{2022b}]{wu2022exploring}
Zhiyuan Wu, Sheng Sun, Yuwei Wang, Min Liu, Quyang Pan, Junbo Zhang, Zeju Li, and Qingxiang Liu.
\newblock Exploring the distributed knowledge congruence in proxy-data-free federated distillation.
\newblock {\em ACM Transactions on Intelligent Systems and Technology}, 2022.

\bibitem[\protect\citeauthoryear{Wu \bgroup \em et al.\egroup }{2024}]{wu2024federated}
Zhiyuan Wu, Tianliu He, Sheng Sun, Yuwei Wang, Min Liu, Bo~Gao, and Xuefeng Jiang.
\newblock Federated class-incremental learning with new-class augmented self-distillation.
\newblock {\em arXiv preprint arXiv:2401.00622}, 2024.

\bibitem[\protect\citeauthoryear{Yang \bgroup \em et al.\egroup }{2020}]{yang2020gradaug}
Taojiannan Yang, Sijie Zhu, and Chen Chen.
\newblock Gradaug: A new regularization method for deep neural networks.
\newblock {\em Advances in Neural Information Processing Systems}, 33:14207--14218, 2020.

\bibitem[\protect\citeauthoryear{Yao \bgroup \em et al.\egroup }{2023}]{yao2021local}
Dezhong Yao, Wanning Pan, Yutong Dai, Yao Wan, Xiaofeng Ding, Chen Yu, Hai Jin, Zheng Xu, and Lichao Sun.
\newblock F ed gkd: Towards heterogeneous federated learning via global knowledge distillation.
\newblock {\em IEEE Transactions on Computers}, 2023.

\bibitem[\protect\citeauthoryear{Zhang \bgroup \em et al.\egroup }{2022a}]{zhang2021fedzkt}
Lan Zhang, Dapeng Wu, and Xiaoyong Yuan.
\newblock Fedzkt: Zero-shot knowledge transfer towards resource-constrained federated learning with heterogeneous on-device models.
\newblock In {\em 2022 IEEE 42nd International Conference on Distributed Computing Systems (ICDCS)}, pages 928--938. IEEE, 2022.

\bibitem[\protect\citeauthoryear{Zhang \bgroup \em et al.\egroup }{2022b}]{zhang2022fine}
Lin Zhang, Li~Shen, Liang Ding, Dacheng Tao, and Ling-Yu Duan.
\newblock Fine-tuning global model via data-free knowledge distillation for non-iid federated learning.
\newblock In {\em Proceedings of the IEEE/CVF Conference on Computer Vision and Pattern Recognition}, pages 10174--10183, 2022.

\bibitem[\protect\citeauthoryear{Zhu \bgroup \em et al.\egroup }{2021}]{zhu2021data}
Zhuangdi Zhu, Junyuan Hong, and Jiayu Zhou.
\newblock Data-free knowledge distillation for heterogeneous federated learning.
\newblock In {\em International Conference on Machine Learning}, pages 12878--12889. PMLR, 2021.

\end{thebibliography}

\end{document}